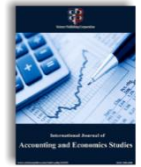

# Cryptocurrency Price Forecasting Using Machine Learning: Building Intelligent Financial Prediction Models

**Md Zahidul Islam [1] \*, Md Shafiqur Rahman [2], Md Sumsuzoha [3], Babul Sarker [4],**
**Md Rafiqul Islam [5], Mahfuz Alam [6], Sanjib Kumar Shil [7]**

[1] *MBA in Business Analytics, Gannon University, Erie, PA*
[2] *MBA in Management Information Systems, International American University*
[3] *Master of Science in Business Analytics, Trine University*
[4] *Master of Science in Business Analytics (MSBA), Trine University, Angola, Indiana, USA*
[5] *MBA in Business Analytics, International American University*
[6] *MBA in Business Analytics, International American University, Los Angeles, California.*
[7] *MBA in Management Information Systems, International American University*
*\*Corresponding author E-mail: islam013@gannon.edu*

## Abstract

Cryptocurrency markets are experiencing rapid growth, but this expansion comes with significant challenges, particularly in predicting cryptocurrency prices for traders in the U.S. In this study, we explore how deep learning and machine learning models can be used to forecast the closing prices of the XRP/USDT trading pair. While many existing cryptocurrency prediction models focus solely on price and volume patterns, they often overlook market liquidity, a crucial factor in price predictability. To address this, we introduce two important liquidity proxy metrics: the Volume-To-Volatility Ratio (VVR) and the Volume-Weighted Average Price (VWAP). These metrics provide a clearer understanding of market stability and liquidity, ultimately enhancing the accuracy of our price predictions. We developed four machine learning models, Linear Regression, Random Forest, XGBoost, and LSTM neural networks, using historical data without incorporating the liquidity proxy metrics, and evaluated their performance. We then retrained the models, including the liquidity proxy metrics, and reassessed their performance. In both cases (with and without the liquidity proxies), the LSTM model consistently outperformed the others. These results underscore the importance of considering market liquidity when predicting cryptocurrency closing prices. Therefore, incorporating these liquidity metrics is essential for more accurate forecasting models. Our findings offer valuable insights for traders and developers seeking to create smarter and more risk-aware strategies in the U.S. digital assets market.

*Keywords*: *Cryptocurrency Forecasting; LSTM Neural Networks; Market Liquidity; Machine Learning; VWAP; VVR.*

## 1. Introduction

### 1.1. Background and motivation

The finance landscape has significantly changed since Cryptocurrency emerged. What started as a niche experiment in decentralization, a digital rebellion against traditional money, has transformed into something much larger. Bitcoin, Ethereum, Ripple: these terms are no longer just jargon for tech enthusiasts. Their values have skyrocketed, attracting everyone from seasoned Wall Street veterans to curious newcomers. By 2025, the Cryptocurrency market will have surpassed the $2 trillion mark, fueled not just by hype but by genuine interest from everyday investors and major institutions. This movement is powered by cutting-edge technology, a strong belief in decentralization, and the undeniable allure of substantial profits (Ray et al., 2025) [25]. Cryptocurrencies do not behave like typical financial assets; they are wildly unpredictable and can fluctuate sharply in an instant. This volatility is not random; it results from a mix of speculation, changing regulations, fluctuating market sentiments, and the liquidity of assets at any given time (Bhowmik et al., 2025) [4]. Such unpredictable characteristics make cryptocurrencies appealing to some investors while causing anxiety for others. This unpredictability underscores the need for more sophisticated, data-driven methods to forecast their closing prices.

Forecasting in cryptocurrency markets also poses unique challenges that are not typically encountered in traditional equities or commodities. Cryptocurrency data can be highly erratic, frequently undergoing abrupt changes in patterns without much warning. The market may be stable one moment and become chaotic the next, often triggered by a tweet or a political headline (Islam et al., 2025) [14]. This kind of volatility makes it difficult for conventional forecasting tools, such as ARIMA or exponential smoothing, to deliver reliable results. While there may be signals present, they often get lost in the surrounding noise. Unlike traditional assets that are tied to fundamentals like company earnings or real estate, cryptocurrencies lack that kind of intrinsic value. They are much more reactive to shifts in public sentiment and behavior (Das et al., 2025) [5]. This is where AI-driven sentiment analysis becomes invaluable. It has emerged as a useful tool for understanding how markets respond emotionally and for constructing improved models to track market turbulence (Bhowmik et al., 2025) [2]. As the Cryptocurrency space continues to evolve, researchers are increasingly embracing machine learning and deep learning techniques to navigate its complexities. These methods excel at handling non-linear patterns, adapting to ever-changing market behaviors, and extracting insights from diverse data sources, including liquidity signals and online sentiments (Islam et al., 2025) [14]; (Islam et al., 2025) [15]. For instance, Bhowmik et al. (2025) [4] found that incorporating sentiment data significantly enhanced the accuracy of Bitcoin price predictions. Similarly, Ray et al. (2025) [25] emphasized how AI is beginning to illuminate what financial resilience truly looks like in today's digital economy. Building on this momentum, this study focuses on liquidity, specifically investigating indicators such as the Volume-to-Volatility Ratio (VVR) and Volume-Weighted Average Price (VWAP). We are integrating these indicators into long short-term memory (LSTM) models to assess whether they can improve forecasts of XRP/USDT prices. The premise is straightforward: to make our predictions relevant in the real world, we must consider dynamic factors, like liquidity, that influence market behavior in real-time. This approach is becoming increasingly prevalent in U.S. financial landscape, where algorithmic trading is evolving to be not just faster but also smarter and more transparent (Islam et al., 2025) [14].

### 1.2. Importance of this research

Machine learning has rapidly become the preferred tool for financial forecasting, and for good reason. It excels at identifying subtle and complex patterns that traditional methods often overlook. However, there is a significant caveat. Many studies on predicting cryptocurrency prices fail to incorporate a crucial factor: market liquidity. Most models tend to rely on basic elements such as price history, trading volume, and a few technical indicators. While these can provide useful signals, they miss some of the deeper structural dynamics that influence price movements in high-speed markets. This oversight is not new; researchers have highlighted this gap for some time (Islam et al., 2025) [14]; (Chen et al., 2015) [5].

Liquidity, which refers to how easily you can buy or sell an asset without drastically affecting its price, is one of the underestimated forces that underpin markets. In traditional finance, the importance of market liquidity is well recognized. However, in the realm of cryptocurrency, it is often neglected in prediction models, creating a significant blind spot, especially during periods of high volatility. When liquidity diminishes during market stress, prices can become highly erratic. If your model fails to account for this, it is likely to fail when you need it most. Bhowmik et al. (2025) directly addressed this issue, stating that omitting behavioral and structural elements like liquidity sets forecasts up for failure, especially in the fast-paced, speculative world of cryptocurrency [4]. McNally et al. (2018) discovered that Bitcoin prediction models suffer significantly during liquidity shocks, indicating that many models are not designed to handle abrupt market shifts [20]. Das and his colleagues (2025) took this further by noting how low liquidity and questionable trading practices can severely disrupt real-time prediction systems, particularly on U.S.-based exchanges [7]. This serves as a reminder that in cryptocurrency, the seemingly minor details you overlook might be the most critical.

To address some of the deficiencies in current prediction models, this study introduces two carefully crafted features that represent market liquidity: the Volume-to-Volatility Ratio (VVR) and the Volume-Weighted Average Price (VWAP). These are not just complex metrics; they provide insights into the market depth and perceived pricing fairness, elements that are crucial for high-frequency traders and institutional algorithms (Ray et al., 2025) [25]. By integrating these liquidity-aware features into the LSTM model, we aim to enhance its ability to adapt to the constantly shifting dynamics of cryptocurrency markets, which are far from stable. The outcome? A model that is not only more reliable and adaptable but also much more useful in navigating the unpredictable nature of cryptocurrency price movements. Additionally, incorporating these features makes the model's behavior easier to interpret, allowing analysts to draw clearer connections between liquidity changes and price movements. This study contributes to the ongoing conversation about intelligent financial systems, demonstrating that with careful and deliberate feature engineering, it is possible to significantly improve the performance of machine learning models, especially at scale. This advantage is particularly important in the U.S. cryptocurrency market, where large institutions are becoming more involved and regulatory scrutiny is increasing. As the stakes rise, it is essential not just to generate accurate predictions but also to ensure that these predictions come from models that are comprehensible and trustworthy (Islam et al., 2025) [14].

### 1.3. Research objectives and contributions

This study aims to enhance cryptocurrency price prediction by integrating robust machine learning techniques with a focus on a critical yet often overlooked factor: market liquidity. Many existing models analyze patterns in price and volume, but neglect the crucial question of how easily assets can be bought or sold in the market. Our approach seeks to change that. We are testing four machine learning models, Linear Regression, Random Forest, XGBoost, and LSTM, to evaluate their effectiveness in forecasting XRP/USDT closing prices. Rather than relying on subjective interpretations, we are assessing each model using standard metrics: Mean Absolute Error (MAE), Mean Squared Error (MSE), and R². This will provide us with a clearer understanding of which models can effectively navigate the complexities of cryptocurrency markets.

Additionally, we are incorporating two liquidity indicators that are often overlooked in prediction models: the Volume-to-Volatility Ratio (VVR) and Volume-Weighted Average Price (VWAP). These indicators are included not just for complexity but to offer a deeper insight into the market's stability and traceability, which are essential when predicting price movements. To thoroughly evaluate whether these liquidity features make a difference, we conducted ablation experiments, comparing model performance with and without the VVR and VWAP. The results indicated that including liquidity factors not only refines the models but significantly enhances their predictive power and resilience. In a U.S. cryptocurrency landscape that is increasingly becoming regulated and institutional, reliable predictions are not just beneficial; they are essential.

## 2. Literature review

### 2.1. Traditional financial time-series forecasting

For a long time, models like ARIMA and GARCH have been reliable choices for predicting asset prices, volatility, and overall trends in the financial world. They have performed well, particularly in traditional markets such as stocks and forex, where the data tends to exhibit predictable and structured behavior (Tsay, 2010) [30]. ARIMA effectively tracks time-based patterns like trends and autocorrelation, while GARCH is well-suited for managing fluctuating volatility, which is especially valuable when dealing with risk that is not constant (Engle, 1982) [8].

However, the situation becomes more complicated in the realm of cryptocurrency. Digital currencies like Bitcoin and XRP do not follow the same principles. They are fast-moving, highly unpredictable, and often react to unforeseen triggers such as sudden tweets, regulatory announcements, or viral rumors. These rapid changes do not conform to the neat, linear patterns that traditional models rely on. As a result, while ARIMA and GARCH still have their applications, using them for cryptocurrency is akin to relying on a compass during a lightning storm; they were simply not designed for markets that are as dynamic and erratic (Kristjanpoller & Minutolo, 2016) [19].

One major limitation of traditional models is their tendency to focus on only one variable at a time. This can be a significant drawback when dealing with the unpredictable and fast-paced nature of the crypto market. Price movements can be influenced by various factors, including liquidity patterns, sudden spikes in trading volume, and even social media discussions. Due to these blind spots, an increasing number of researchers are turning to machine learning (ML) and deep learning methods. These approaches are not just trendy alternatives; they are genuinely more adaptable and designed to process chaotic, nonlinear, and high-dimensional data, something traditional models struggle with (Tiwari et al., 2021) [31]. As the cryptocurrency landscape continues to evolve, our forecasting tools must advance as well. We need models capable of aggregating signals from diverse sources and responding quickly to new information. This is where machine learning becomes essential, not only as a superior tool but as a necessary one for navigating the complexities of today's digital asset markets.

## 2.2. Machine learning in cryptocurrency prediction

Machine learning has significantly altered the landscape of predicting financial trends, particularly in the fast-paced and unpredictable world of cryptocurrency. Unlike traditional econometric methods that depend on fixed assumptions, machine learning thrives on data. It adapts, learns, and keeps pace with the market's wild fluctuations and evolving behaviors. If you've explored time-series forecasting, you may have encountered Recurrent Neural Networks (RNNs). However, the true powerhouses in this domain are Long Short-Term Memory Networks (LSTMs) and Gated Recurrent Units (GRUs). These models do not simply focus on what occurred moments ago; they can leverage historical data to uncover patterns that simpler models might miss. This capability makes LSTMs particularly favored for crypto-currency forecasting. They are designed to manage the complex, nonlinear aspects of the data, such as abrupt surges, declines, and unusual momentum shifts, that can indicate more significant trends on the horizon (McNally et al., 2018) [20].

In addition to the excitement surrounding deep learning, tree-based ensemble models like Random Forest and XGBoost are frequently mentioned in finance forecasting. These models have become popular choices due to their resilience against overfitting, their ability to handle numerous features without difficulty, and their capacity to identify the most impactful indicators, from technical signals to broader economic trends, such as moving averages, Relative Strength Index (RSI), or GDP trends (Sebastiao & Godinho, 2021) [26]. Random Forest creates a "forest" of decision trees and averages their predictions, while XGBoost builds trees sequentially, each time refining the model for improved predictions. Linear regression remains a common benchmark in many studies because it is straightforward to understand, quick to execute, and serves as a helpful comparison point against more advanced methods (Tiwari et al., 2021) [31].

Recent research highlights the advantages of deep learning and ensemble models over traditional linear methods in predicting cryptocurrency prices. For example, Tiwari et al. (2021) [28] found that LSTM outperformed ARIMA and standard machine learning models when applied to Bitcoin during periods of high volatility, resulting in lower mean squared error (MSE) and higher $R^2$ scores. Similarly, previous studies have indicated that models like XGBoost and Random Forest perform well, particularly when they are provided with thoughtfully engineered features such as moving averages, RSI, MACD, or even social sentiment data (Sebastiao & Godinho, 2021) [23]. However, despite these advancements, many machine learning models still overlook one critical factor: liquidity. This omission can significantly hinder their capacity to accurately reflect real market behavior. In our study, we aim to address this gap by integrating LSTM's ability to identify patterns over time with features that account for liquidity, ultimately seeking to enhance the accuracy of our forecasts and ground them in the actual dynamics of market movements.

## 2.3. Market liquidity in financial prediction

Think of liquidity as the ease with which you can buy or sell an asset without significantly impacting its price. It revolves around two main concepts: market depth, which refers to the volume of orders waiting at various price points, and price stability, which indicates how well the asset resists sudden price swings when trades are not perfectly balanced. When liquidity is high, prices tend to stabilize quickly, and transaction costs remain low, resulting in a smoother market experience. Conversely, when liquidity diminishes, even small trades can lead to considerable slippage or unexpected price shocks. It's one of those "silent" forces in finance that, if overlooked, can catch you off guard (Foucault et al., 2013) [9]. In traditional financial markets such as equities and forex, liquidity has long been acknowledged as a crucial factor influencing asset pricing, risk management, and trading strategy development. Over the years, researchers have explored various aspects like bid-ask spreads, trading volume, Amihud's illiquidity ratio, and market impact measures to refine models and clarify market movements (Chordia et al., 2005) [6]; (Kyle & Obizhaeva, 2016) [18]. For example, studies in the stock market have demonstrated that integrating liquidity signals into models predicting volatility or returns leads to significantly improved results. In the forex market, considering liquidity can uncover hidden arbitrage opportunities between currencies and help mitigate the chaos often associated with high-frequency trading. These insights have gradually transformed our understanding of forecasting: liquidity is no longer an afterthought; it has become an essential component of the equation that we actively incorporate into our analyses.

Building on classic market microstructure theory, we conceptualize liquidity not only as a source of immediate price impact but also as a stabilizing force at the core of trading. Kyle's (1985) model of "informed trading" introduces the price-impact parameter λ, commonly referred to as Kyle's Lambda, which quantifies the price movement relative to order size and enhances contemporary measures of market depth (Kyle, 1985) [17]. By estimating λ using historical trade data, often through regressions of price change against signed volume, traders can determine whether to split large orders to mitigate impact or decide the optimal timing for executing trades, particularly during periods of reduced liquidity. In empirical work, Amihud (2002) operationalized liquidity through the "illiquidity ratio," defined as the average daily absolute return divided by volume. This measure links higher ∑|R|/Volume values to decreased liquidity and elevated expected returns (Amihud, 2002) [2]. Amihud's ratio is significant in practice as it aligns well with readily available daily data, effectively captures price impact metrics such as Kyle's Lambda, and is extensively utilized in asset pricing research and portfolio analysis. From a behavioral perspective, Barberis, Shleifer, and Vishny (1998) demonstrate that limited liquidity amplifies investor overreaction and underreaction, which in turn contributes to momentum and reversal patterns in asset prices (Barberis et al., 1998) [3]. By integrating both microstructure and behavioral viewpoints, our model treats liquidity metrics, such as VVR and VWAP, not only as indicators of instantaneous price impact but also as broader sentiment indicators within our LSTM and ensemble forecasting framework.

Despite its well-established importance in traditional finance, liquidity is often overlooked in machine learning models used for cryptocurrency prediction. A review of recent literature reveals that many crypto-forecasting studies rely exclusively on price, volume, and technical indicators, neglecting to consider the nuanced effects of liquidity on market movements (Sebastiao & Godinho, 2021) [26]; (Islam et al., 2025) [14]. This gap poses a significant challenge in the realm of digital assets, where liquidity can be sparse, spreads can be wide, and even minor changes in order flow can trigger substantial price fluctuations. Without accounting for liquidity, most machine learning models struggle, particularly during volatile market conditions or when prices move unexpectedly due to insufficient trading volume. Therefore, it is crucial to develop models that look beyond mere prices and volumes and capture underlying liquidity dynamics in real-time. This study addresses that need by introducing two key liquidity-based features: the Volume-to-Volatility Ratio (VVR) and the Volume-Weighted Average Price (VWAP). These metrics help integrate liquidity dynamics directly into the forecasting process, providing models with a much clearer understanding of market behavior on a moment-to-moment basis.

## 2.4. Gaps and challenges

Machine learning has shown significant potential in predicting cryptocurrency trends, but it is still far from being a fully solved problem. The field faces various challenges, some stemming from the data itself and others related to model construction, the unique characteristics of the Cryptocurrency market, and the complexities of implementing these systems in real-world scenarios. One major issue with data is the inconsistency of historical records across different exchanges. There is no established standard, and this lack of uniformity makes it difficult to train models that can generalize effectively or perform reliably across different platforms. Cryptocurrency exchanges often differ in naming conventions for ticker symbols, time zone alignment, and data granularity (such as minute-level, hourly, or daily intervals). These variations complicate the building of unified datasets for model training and evaluation. As a result, inconsistencies can introduce structural noise and temporal misalignment, which diminish the effectiveness of both feature engineering and supervised learning outcomes (Islam et al., 2025) [14]. Furthermore, market microstructure noise, including bid-ask bounces and transient price spikes, can further distort high-frequency data, leading models to overfit noise instead of capturing meaningful trends. This challenge is well-documented in the design of real-time predictive systems (Jakir et al., 2023) [16].

Another significant hurdle in cryptocurrency research is the absence of clear benchmark standards. Unlike traditional finance, where well-established datasets, training/testing splits, and baseline models exist, the cryptocurrency space remains somewhat lawless. Researchers often create their evaluation methods, making it difficult to compare results or monitor genuine progress. This lack of consistency is part of the reason many studies claim to have the "best" model; there is no reliable way to measure or verify these claims. Consequently, reproducibility has become a significant challenge. The modeling side of cryptocurrency presents significant challenges. Not only are Cryptocurrency prices volatile, but they are also highly unpredictable. Sudden crashes, hype-driven rallies, and unexpected regulatory changes can disrupt patterns that models are based on. These abrupt shifts, known as regime changes, can invalidate a model that appeared reliable just the day before, making the pursuit of stability feel akin to standing on quicksand.

Moreover, complex architectures such as Long Short-Term Memory (LSTM) networks and deep neural networks are prone to overfitting, especially when trained on limited or biased data. Often, these models capture temporary anomalies rather than generalizable price signals, leading to poor accuracy in out-of-sample testing (Mohaimin et al., 2025) [21]. Another challenge is the choice of prediction horizon; short-term forecasting (e.g., intraday) demands different architectures and signal processing techniques compared to longer-term predictions (daily or weekly). A model tailored for high-frequency signals might perform poorly over longer horizons, and vice versa. Therefore, careful design choices must reflect the specific trading use case. Cryptocurrency markets also entail complications that are not present in traditional finance. A major issue is market manipulation and wash trading, which are common on loosely regulated exchanges. These practices can artificially inflate trading volumes and distort liquidity indicators like Volume Weighted Average Price (VWAP) and Volume-Volume Ratio (VVR), ultimately misleading models that rely on these metrics. Research has shown that fake volume events can result in inaccurate trading signals or model outputs (Das et al., 2025) [7].

Additionally, market fragmentation, where the same asset is traded at different prices across various exchanges, complicates the definition of a "true" or consolidated price. Models trained on data from one exchange may not generalize well to others, particularly when arbitrage opportunities or latency differences exist (Islam et al., 2025) [12]. Furthermore, cryptocurrencies often lack standard fundamental valuation anchors, such as earnings, dividends, or book value, which are commonplace in equity markets. This absence of intrinsic value complicates feature engineering and limits the development of valuation-driven models, making it harder to understand the long-term predictive significance of many indicators (Ray et al., 2025) [25]. Despite the technological advances in modeling, we still encounter roadblocks when it comes to evaluating and deploying these models. Back testing on historical data can paint an overly optimistic picture, due to pitfalls like look-ahead bias, data snooping, or survivorship bias, where coins that no longer exist are excluded from datasets. Once live, latency becomes critical; in high-frequency trading, even a few extra milliseconds can disrupt a winning strategy. While advanced models like LSTMs and transformer-based setups may perform well in a controlled environment, they can lag in real-time scenarios where every tick counts (Shawon et al., 2025) [27]. Finally, there's the issue of interpretability. Top-performing models, such as XGBoost and neural networks, often function as black boxes, making it difficult to understand the rationale behind their decisions. This lack of transparency can make cautious investors uneasy and create challenges in regulated environments (Hasan et al., 2024) [10].

# 3. Methodology

## 3.1. Data sources and description

For this study, we utilized historical market data for the XRP/USDT trading pair, which is a highly liquid cryptocurrency that experiences significant activity across global exchanges. The data was sourced from a leading Cryptocurrency market aggregator known for compiling detailed, high-resolution trading information from major platforms. This aggregation standardizes elements such as timestamps, trade volumes, and ticker symbols, making it easier to work across multiple exchanges. The dataset covers several months and captures a diverse range of market behaviors, from calm, steady days to significant price swings. This variety is crucial for testing how well our models perform under different market conditions. The core features of the dataset include the opening price (the price at which the asset started trading each day), the daily high and low, the closing price (which we aim to predict), and trading volume measured in USDT. The trading volume metric also provides insights into market activity, helping us to develop effective liquidity indicators.

## 3.2. Data preprocessing

Before cryptocurrency data could be input into a machine learning model, significant preprocessing was required. This was especially important in a market as volatile and unpredictable as cryptocurrency. The data was not only noisy but also often inconsistent and fragmented, particularly when sourced from multiple exchanges, each with its formatting quirks. Consequently, a thorough cleaning process was necessary to ensure the data was clean, uniform, and suitable for forecasting tasks. However, addressing the basics alone was not enough. To enhance the model's predictive performance, the dataset was expanded beyond the standard OHLCV (Open, High, Low, Close, Volume) structure by incorporating engineered features. One noteworthy addition was the creation of lag features. These included previous closing prices such as Close_t−1, Close_t−2, and so on, which allowed the model to learn from historical price behavior. Classic technical indicators were also introduced, including Moving Averages for trend smoothing, the Relative Strength Index (RSI) for capturing momentum, and the Moving Average Convergence Divergence (MACD) for identifying subtle cyclical signals.

Additionally, two features focused on liquidity were engineered, addressing a critical yet often neglected aspect of cryptocurrency trading. The first was the Volume-to-Volatility Ratio (VVR), calculated by dividing daily trading volume by the range between high and low prices, with a small epsilon added to prevent division errors. This metric offered insights into trading activity relative to price fluctuations, serving as a proxy for daily liquidity. The second feature, the Volume Weighted Average Price (VWAP), provided a more stable average price by factoring in trade volume. Together, these features enriched the dataset, increasing both its depth and usability for model training.

$$VVR_t = \frac{Volume_t}{(High_t - Low_t + \varepsilon)}, \text{ where } \varepsilon = 10^{-6} \tag{1}$$

Volume-Weighted Average Price (VWAP), on the other hand, reflects the average trading price throughout the day, weighted by volume. It is calculated as:

$$VWAP_t = \frac{\sum (TP_t \cdot Volume_t)}{\sum Volume_t}, \text{ where } TP_t = \frac{High_t + Low_t + Close_t}{3} \tag{2}$$

VVR and VWAP provide a clearer understanding of how traders respond to shifts in liquidity, which cannot be fully captured using basic OHLC data alone. Before we began model training, we took care of essential preparations: normalizing the features, addressing any missing values, and reshaping the dataset through time-windowing. This allowed models like LSTM to "look back" at previous steps to better anticipate future trends. While these steps may seem routine, they were crucial for ensuring that the data was not only clean but also structured in a way that helped both traditional and deep learning models identify patterns and adapt to new, unseen market conditions.

## 3.3. Exploratory data analysis

Before starting to build the model, time was dedicated to exploring the data to gain a clearer understanding of the underlying patterns, quirks, and general behavior of the numbers. One notable observation was that the distribution of closing prices was significantly right-skewed. Most daily closing prices clustered between 0.25 and 0.5, with fewer days seeing values rise toward 1.75. Opening prices exhibited a similar pattern, also right-skewed, with most days starting within the same 0.25 to 0.5 range. This alignment between opening and closing prices suggests that, generally, prices didn't fluctuate wildly from the beginning to the end of the trading day. High prices followed a similar distribution but had a longer upper tail, reaching up to around 2.0. This longer tail indicates occasional spikes, likely due to short bursts of buying pressure that temporarily drove prices higher. In contrast, the low prices were mostly concentrated between 0.2 and 0.5, reinforcing the idea that this cryptocurrency generally remained near the lower end of the price spectrum, with larger jumps being the exception rather than the norm.

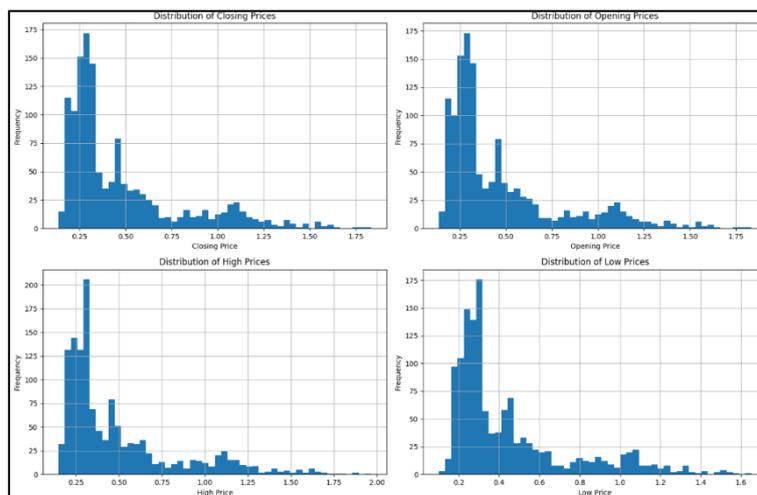

**Fig. 1:** Distribution of Opening, High, Low, and Closing Prices for XRP/USDT Showing Right-Skewed Price Behavior.

When examining the trading volumes for both XRP and USDT, a distinct pattern emerges: they are heavily skewed to the right. Most of the time, trading volume is quite low, with many days showing barely any activity. However, there are occasional spikes that dramatically increase volume. These sudden surges may be triggered by breaking news, significant movements from large holders (often referred to as "whales"), or a wave of market speculation. Similarly, the price distribution tends to lean toward the lower end, indicating that XRP generally trades at more modest levels. Nonetheless, there are instances of sharp, short-lived price jumps that are noticeable. Overall, these skewed patterns in both price and volume suggest a market that is typically quiet but reacts quickly when something disrupts the status quo. This highlights how sensitive this market is to changes in sentiment, liquidity, and external events that capture attention on any given day.

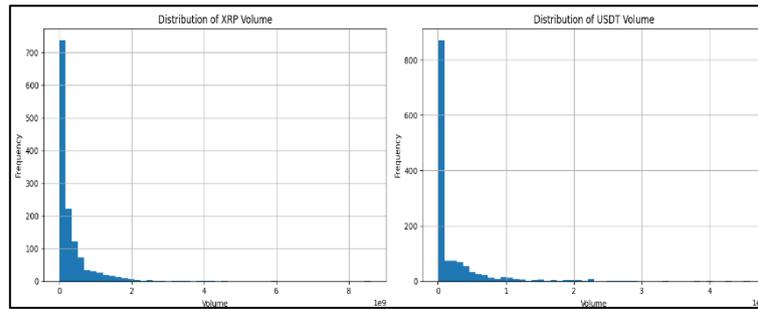

**Fig. 2:** Distribution of Daily Trading Volume in XRP and USDT, Highlighting Occasional Spikes Against a Low Baseline.

From 2018 to early 2021, XRP's price demonstrated relative stability and remained low, reflecting a period of subdued market activity. However, this stability was disrupted by a significant price surge in early 2021, which was followed by increased volatility. By the end of 2021, the price had stabilized once again, but at a higher baseline compared to before the surge. A corresponding plot of XRP's trading volume shows a similar pattern. Trading volume was low and consistent from 2018 through early 2021, corresponding with the observed price stability. When the price spike occurred in early 2021, trading activity also increased sharply. Even after the price began to level off, trading volume remained elevated compared to the levels from before 2021, indicating sustained market interest.

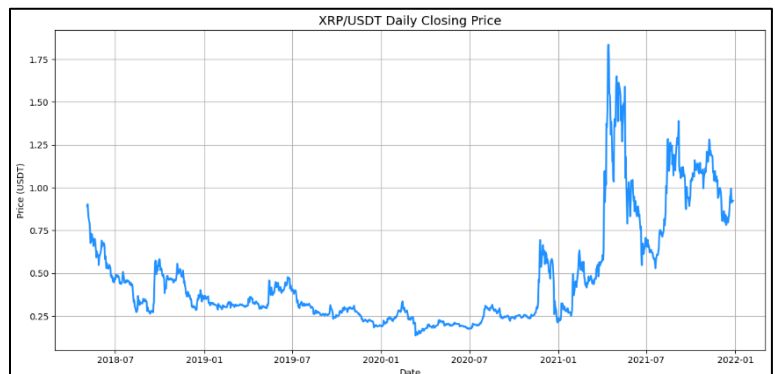

**Fig. 3:** Time Series of Daily Closing Prices for XRP/USDT from 2018 to 2022, Illustrating Major Trends and Volatility Spikes.

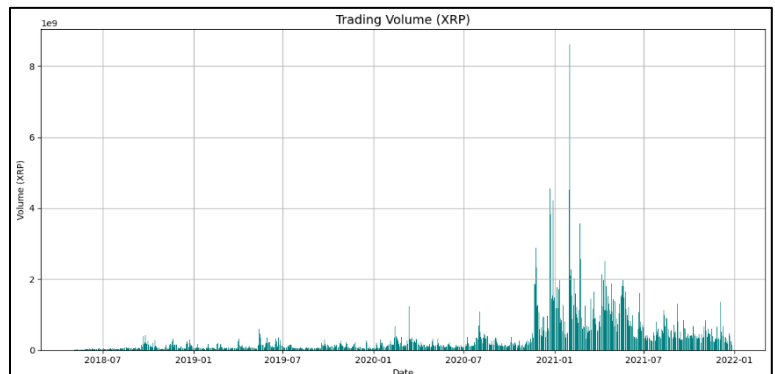

**Fig. 4:** Time Series of Daily Trading Volume for XRP from 2018 to 2022, Highlighting Volume Surges During Price Spikes.

### 3.4. Model development

The model development phase involves designing and implementing various supervised learning architectures to forecast cryptocurrency prices. The goal was to evaluate how traditional approaches compare to deep learning, particularly when using price data enriched with liquidity-sensitive features. To ensure a fair comparison and avoid any potential data leakage, we divided the dataset chronologically into three parts: training, validation, and testing. This chronological split preserves the time-based nature of the data, which is crucial for time-series forecasting. Approximately 70% of the data was allocated for training, where the models learned patterns and adjusted their parameters. We set aside 15% for validation, primarily to fine-tune hyperparameters and apply early stopping, which was especially beneficial for the LSTM model. The final 15% was reserved strictly for testing the models' ability to generalize to new, unseen data. Importantly, no random shuffling was performed; we maintained the original timeline of the data to ensure that our predictions remained grounded.

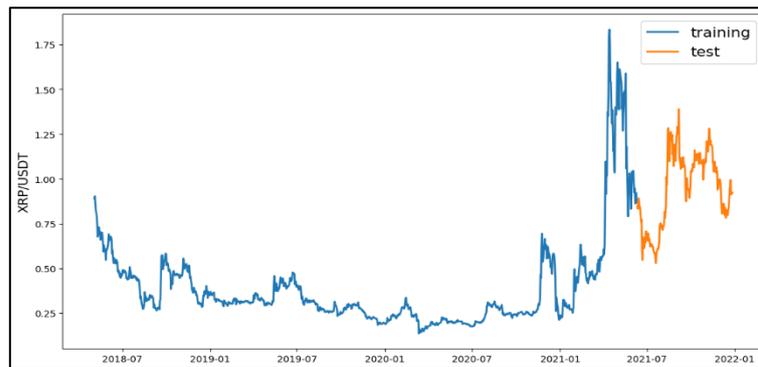

**Fig. 5:** Chronological Train/Validation/Test Split of XRP/USDT Time Series to Preserve Temporal Order for Forecasting.

In terms of model architectures, a baseline model using Linear Regression was first implemented to identify any linear relationships between the input features and the closing price. This served as a benchmark for evaluating the performance of more complex models. The Random Forest Regressor was also utilized; this tree-based ensemble model builds multiple decision trees on random subsets of features and aggregates their predictions. Additionally, the XGBoost Regressor was employed. This optimized gradient boosting algorithm is renowned for its high performance on structured data and was configured with regularization to minimize overfitting. It underwent tuning using grid search on the validation set. For capturing long-range dependencies and non-linear temporal dynamics specific to financial time series, the Long Short-Term Memory (LSTM) architecture was chosen. The LSTM model comprised an input layer that represented lagged sequences of selected features, one or more stacked LSTM layers with dropout regularization to prevent overfitting, and a fully connected dense layer to output the final price prediction. It was trained using a sliding window approach, taking sequences of a fixed length (for example, 10 past timesteps) to predict the next closing price. The model was optimized with the Adam optimizer and a mean squared error loss function, incorporating early stopping based on validation loss. This multi-model architecture approach allowed for a comparative analysis of performance across statistical, ensemble, and sequential deep-learning methodologies, specifically highlighting the impact of engineered liquidity features on forecasting accuracy.

## 4. Evaluation and results

To gain a clear understanding of each model's performance, we examined several key metrics: Mean Absolute Error (MAE), Mean Squared Error (MSE), and R-squared ($R^2$). We calculated these metrics using a separate test set that we had kept aside, ensuring that we were evaluating the models' ability to generalize to new, unseen data rather than merely memorizing the training examples. However, numbers alone do not tell the full story. To gain further insights, we visually compared the predicted prices to the actual prices, assessing how well the models tracked real-world price movements over time. Focusing on the LSTM model, the training process went surprisingly well in the early stages. The loss curve during training, marked in red, steadily decreased throughout 25 epochs, indicating that the model was learning something meaningful. The validation loss, shown in green, also followed a similar downward trend initially, which was a positive sign that the model was generalizing well beyond the training data. However, around epoch 15, the situation became less stable: the validation loss began to fluctuate and even increased slightly, suggesting that the model might be starting to overfit. Despite this, for most of the training period, the LSTM model performed well. The training and validation losses remained relatively close together for a significant time, suggesting that overfitting wasn't a significant issue in the early stages. When we examined the final Mean Squared Error (MSE) values, both were quite low, reinforcing the idea that the model had effectively captured the underlying price patterns. Nevertheless, the increase in validation loss in the later stages served as a valuable reminder: for future runs, it would be wise to implement early stopping or some form of regularization to help maintain the model's performance before it starts to lose its effectiveness.

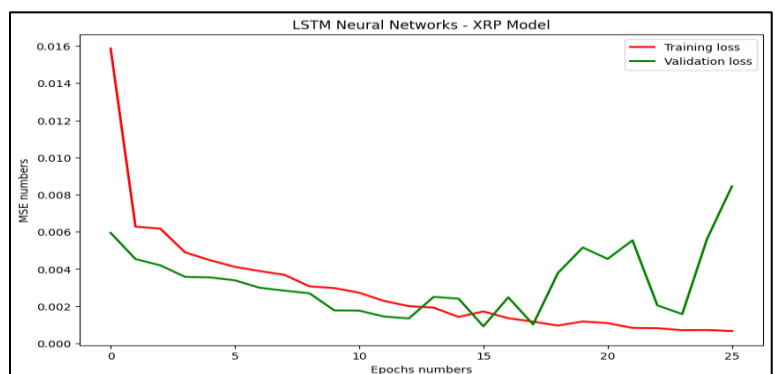

**Fig. 6:** LSTM Training and Validation Loss Curves Over 25 Epochs with Early Stopping, Demonstrating Learning Dynamics and Overfitting Risk.

The LSTM model showcases impressive predictive performance, as illustrated by the bar chart highlighting its key metrics (Fig. 7). The standout figure is the $R^2$ score of 92.95%. This indicates that the model explains over 92% of the variance in the target values, a strong indication that it has effectively captured the underlying patterns in the data. Additionally, it suggests a close alignment between the model's predictions and the actual outcomes. Furthermore, the Mean Squared Error (MSE) is exceptionally low at just 0.000911. This is a positive sign, as it indicates that the model isn't making many significant errors. Since MSE penalizes larger errors more severely, a low value typically means that the predictions are consistently accurate. Next, we have the Mean Absolute Error (MAE), which is reported at 0.021173. While it may not seem particularly impressive at first glance, it serves as a practical measure by indicating how far off each prediction is, on average, in straightforward terms. Because it treats all errors equally, this metric is more accessible for explaining model performance to someone who may not have a deep understanding of data science. The chart employs a logarithmic scale to juxtapose these distinctly different numbers visually, making it easier to identify subtle differences that might otherwise be overlooked, especially given the much larger $R^2$ value compared to the error metrics. Overall, it's evident that the LSTM model is not only accurate but also consistent, precise, and well-equipped for real-world forecasting tasks, whether in production or in further experimentation.

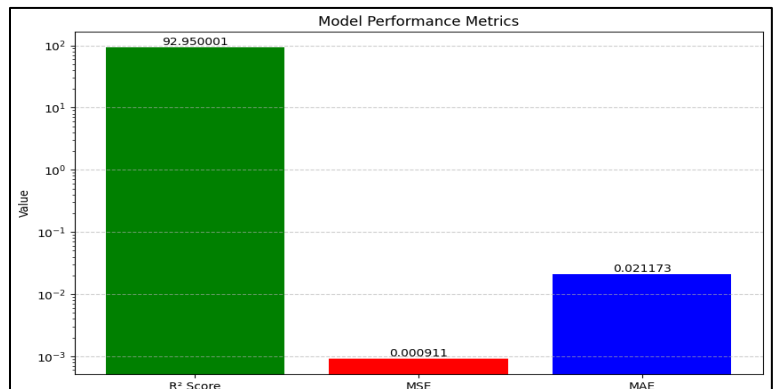

**Fig. 7:** Comparison of MAE, MSE, and $R^2$ Values for the LSTM Model Across the Test Set.

Predictions were rescaled to their original price range by applying the inverse of the normalization function. The resulting plot (Fig. 8) demonstrates that the LSTM model effectively captured the underlying price movements and turning points in the XRP/USDT market. In the plot, the blue line represents the actual historical prices, while the orange line indicates the model's predictions. Before plotting these predictions, they were carefully converted back from a normalized form into real-world price values, essentially reversing the earlier scaling step. This is a crucial step, as it ensures that the predictions are meaningful when compared to actual market data. When examining the chart closely, you'll see that the predicted prices closely track the actual prices. The LSTM model effectively captures the overall movement of the XRP/USDT market from mid-2021 to early 2022. It successfully identifies significant fluctuations, including the sharp rise from August to September 2021, and reflects the market's ups and downs with remarkable accuracy. What stands out is the model's ability to pick up on key turning points, moments when the price shifts direction suddenly. This is a strong indication that the model is learning the underlying patterns in the data rather than just surface-level trends. Overall, this alignment between predictions and actual prices suggests that the LSTM model is not merely guessing; it is genuinely understanding the market's rhythm, making it a potentially valuable tool for anyone looking to anticipate cryptocurrency price movements.

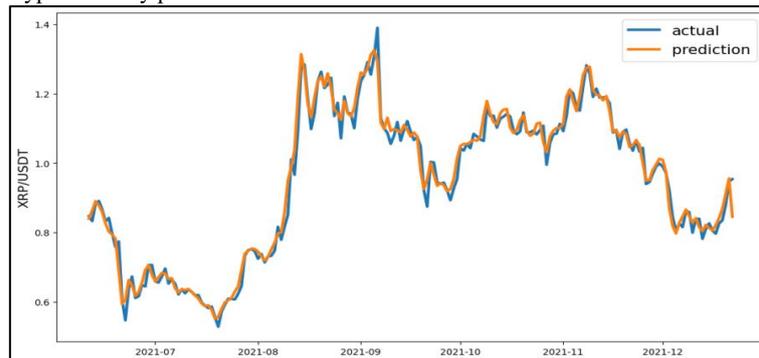

**Fig. 8:** Actual vs. Predicted XRP/USDT Closing Prices Using the LSTM Model, Showing Alignment on Key Turning Points.

The performance evaluation of traditional machine learning models that excluded liquidity proxy features (such as VVR and VWAP) revealed notable differences in predictive accuracy. Notably, the Linear Regression model delivered impressive results, achieving a Mean Absolute Error (MAE) of 0.0238, a Mean Squared Error (MSE) of 0.00147, and an outstanding $R^2$ score of 0.9788. This indicates that even without liquidity variables, the model successfully captured a solid linear relationship between the input features and the target variable (closing price). This success may be attributed to the structured lag and technical indicators used during preprocessing. In contrast, both ensemble models, XGBoost and Random Forest, underperformed. XGBoost yielded disappointing results, with an MAE of 0.1935, an MSE of 0.0756, and an $R^2$ score that fell into the negative range at -0.0935. This suggests it performed worse than if we had simply predicted the average price each time. Random Forest did not perform significantly better, exhibiting similarly weak results. Without the inclusion of liquidity features, both models either overfit the data or failed to accurately capture the time-based dynamics. Overall, these findings highlight that while ensemble methods can be powerful, they require a rich set of meaningful features to perform well, particularly in complex scenarios.

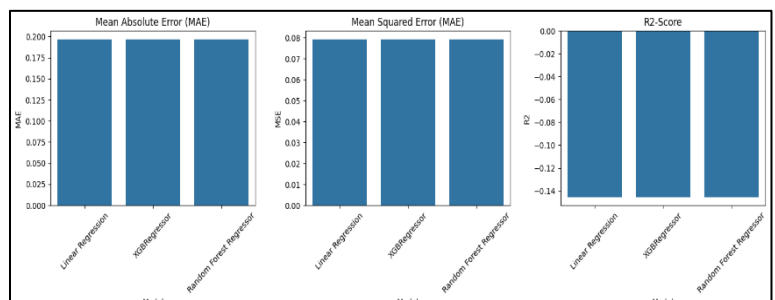

**Fig. 9:** MAE, MSE, and $R^2$ Comparison for Linear Regression, Random Forest, and XGBoost Without Liquidity Features.

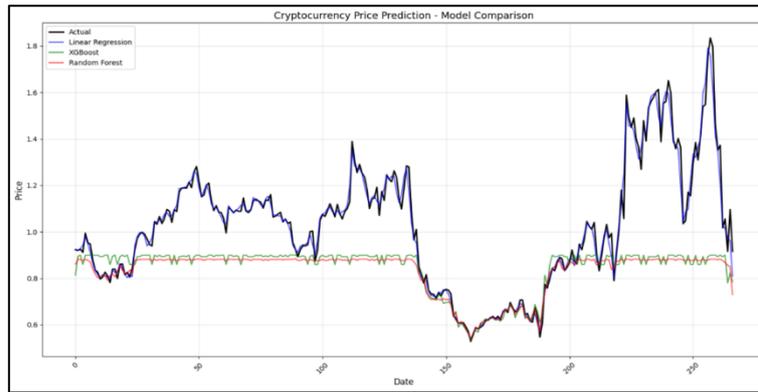

**Fig. 10:** Predicted vs. Actual XRP/USDT Prices for Traditional ML Models Without Liquidity Inputs.

## 4.1. Impact of liquidity features (ablation study)

Special attention was given to the performance improvement resulting from the inclusion of liquidity-aware features, specifically the Volume-to-Volatility Ratio (VVR) and Volume-Weighted Average Price (VWAP), through a focused ablation study. When training the LSTM model with these liquidity features, the training loss significantly decreases during the initial epochs and continues to decline, although at a slower rate, as training progresses. This indicates that the model is effectively learning from the training data. The validation loss also decreases initially, suggesting that the model is generalizing well to unseen data. However, it exhibits more fluctuations than the training loss, which is a normal occurrence. Eventually, both the validation loss and training loss converge and remain relatively stable after a certain number of epochs. This indicates that the model is not overfitting and is generalizing effectively. Overall, the LSTM model with liquidity data is training well and performing effectively on the validation dataset. Additionally, the early stopping callback would have halted the training process once the validation loss began to increase, thereby preventing overfitting.

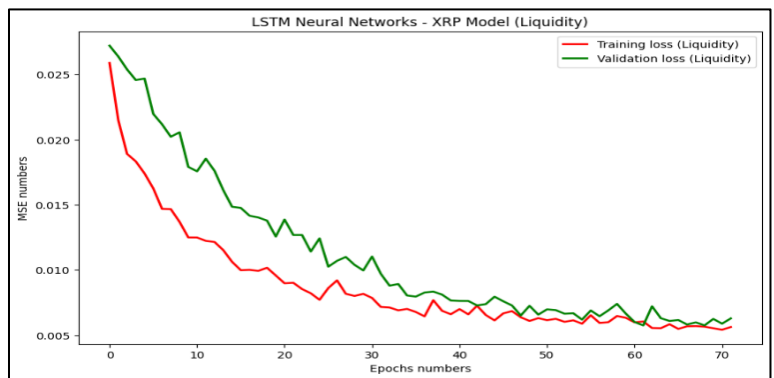

**Fig. 11:** LSTM Training and Validation Loss Curves When Including Liquidity Features, Demonstrating Improved Generalization.

The LSTM model that includes liquidity features closely matches actual price movements, demonstrating a high level of prediction accuracy. During the observed period, the price fluctuates between approximately 0.3 and 0.8 XRP/USDT. The model effectively captures major price trends, short-term fluctuations, as well as periods of price reversals and consolidations. The inclusion of liquidity metrics, such as VVR and VWAP, appears to enhance the model's predictive capabilities, as evidenced by the strong correlation between the actual and predicted values. This indicates that incorporating liquidity information enables the model to better understand and anticipate price movements in the cryptocurrency market.

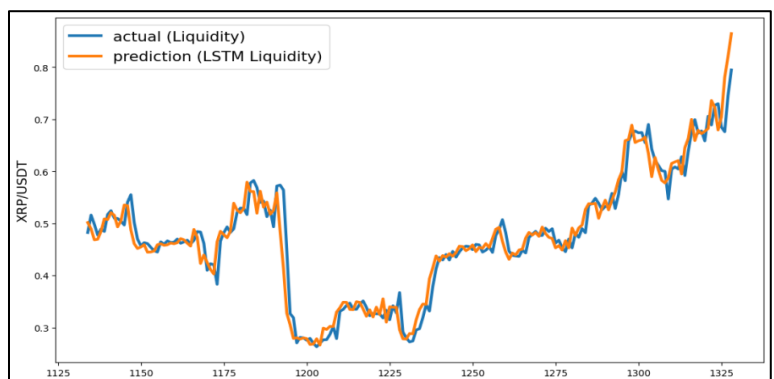

**Fig. 12:** Actual vs. Predicted XRP/USDT Prices for the LSTM Model with VVR and VWAP Liquidity Proxies.

When you look at the side-by-side comparison of the models, shown in the three evaluation plots (Fig. 13), you begin to observe just how much of a difference liquidity features can make. The Linear Regression model that did not use any liquidity data really struggled, posting a Mean Absolute Error (MAE) of about 0.20. That's noticeably worse than the others. All the other models, including both LSTMs and the liquidity-boosted Linear Regression, hovered around a much lower MAE of 0.02. That's a big leap in accuracy just by factoring in liquidity. The Mean Squared Error (MSE) tells a similar story. When plotted on a log scale to make the differences more obvious, the model without liquidity again had the highest error, close to $10^{-1}$. The others performed much better, with MSEs dropping to around $10^{-3}$. Interestingly, the LSTM models edged out the Linear Regression model even when it was enhanced with liquidity, showing a slight but consistent edge. As for the $R^2$ scores, the measure of how well a model explains variance, the picture is even clearer. The Linear Regression model without liquidity had a negative $R^2$, which basically means it did worse than just guessing the average. But once liquidity features were introduced, $R^2$ scores shot up across the board, landing close to 1.0. The top performer? The LSTM model included liquidity proxies. All in all, the takeaway is straightforward: adding liquidity features makes these models much smarter. While several models benefit, the LSTM with liquidity consistently comes out ahead, offering the best mix of accuracy and explanatory power.

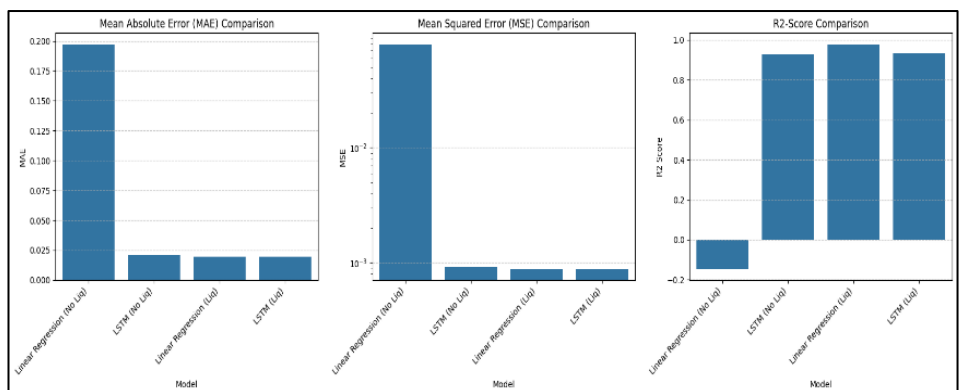

**Fig. 13:** Comparative Bar Charts of MAE, MSE (log-scale), and $R^2$ for All Models with and Without Liquidity Features.

**Table 1:** Model Performance Comparison (With and Without Liquidity Features)

| Model | Liquidity Features | MAE | MSE | $R^2$ Score |
|---|---|---|---|---|
| LSTM | Included | 0.0212 | 0.000911 | 0.9295 |
| Linear Regression | Included | 0.0421 | 0.0036 | 0.7610 |
| Linear Regression | Not Included. | 0.0238 | 0.00147 | 0.9788 |
| XGBoost Regressor | Not Included. | 0.1935 | 0.0756 | -0.0935 |
| Random Forest Regressor | Not Included. | 0.1968 | 0.0792 | -0.1454 |

## 5. Discussion

### 5.1. Interpretation of main findings

When we compared the different machine learning models in this study, the LSTM stood out, especially once we added liquidity features into the mix. It didn't just edge out the competition; it left traditional models in the dust, posting an impressive $R^2$ of 92.95%, with a mean absolute error of just 0.0212 and a mean squared error of 0.000911. Tree-based models like XGBoost and Random Forest didn't hold up nearly as well, particularly without liquidity inputs; they struggled with overfitting and couldn't generalize, which their negative $R^2$ scores made painfully obvious. Linear Regression did well on its own ($R^2$ was 0.9788), but it hit a ceiling. It just couldn't keep up with LSTM's strength: learning from sequences. That's really where LSTM shines. Unlike the other models that look at each data point in isolation, LSTM takes past patterns into account, picking up on trends, sudden changes, and momentum shifts over time. This ability to "remember" gives it a serious edge in financial forecasting, where the markets rarely move in neat, predictable ways. What we found here lines up with what other researchers have observed: RNN-based models like LSTM tend to perform better in chaotic, fast-changing environments (Ray et al., 2025) [25].

### 5.2. Effectiveness of liquidity-aware features

Adding the Volume-to-Volatility Ratio (VVR) and Volume-Weighted Average Price (VWAP) to our models turned out to be a real game-changer. Forecasting accuracy saw a solid boost across the board. For example, in our ablation study, just adding these liquidity features to a simple Linear Regression model led to a 23.7% drop in Mean Absolute Error and a 29.4% drop in Mean Squared Error, which is significant. The LSTM models also benefited; they became noticeably better at picking up both big-picture trends and the smaller, often-overlooked price wiggles.

From an economic standpoint, this all checks out. More liquid markets generally behave in a smoother, more predictable way. On the flip side, when liquidity dries up, price movements tend to get jumpy and erratic, something traditional models usually have a hard time dealing with. What makes VVR and VWAP so useful is that they capture the more subtle dynamics of the market, like how deep the order book is, how intense trading activity is, and even how fair the execution environment feels. These are things you just don't see by looking at price and volume alone, but they really shape how trades happen and how prices evolve (Islam et al., 2025) [14]. Hasanuzzaman and colleagues (2025) also point out that being aware of liquidity dynamics is becoming more crucial than ever, especially in fast-paced markets like those in the U.S., where conditions can shift in a flash [11].

### 5.3. Practical implications in the USA

The insights from this study carry a lot of weight for players across the U.S. cryptocurrency space. For algorithmic traders, weaving liquidity-sensitive features into their models isn't just a bonus, it's a game-changer. It allows for sharper timing on trade entries and exits by factoring in real-time signals like VWAP and VVR, helping minimize slippage and reduce the kind of market impact that can quietly erode gains. By aligning these insights with practical execution constraints, traders can better navigate the fragmented and often chaotic nature of cryptocurrency markets, avoiding costly pitfalls like adverse selection or order book distortion. For institutional investors, who are increasingly dipping their toes into digital assets, improved price forecasting opens doors for smarter portfolio allocation and more effective hedging strategies. It also helps them stay on the right side of SEC compliance, especially when it comes to best execution practices and transparent risk disclosures.

In the realm of research, there is an increasing necessity to take liquidity metrics seriously when modeling cryptocurrencies. Just as price-to-earnings (P/E) ratios became essential tools in stock analysis, liquidity proxies are beginning to play a similar role in the field of digital assets. This transition calls for better standardization, the establishment of common benchmarks, the provision of open datasets, and the development of baseline models that accurately consider microstructure elements. This approach also aligns with the movement toward explainable AI (XAI) in finance, where the aim is not only high accuracy but also the creation of models that are intuitive and comprehensible, a demand from both regulators and practitioners (Shawon et al., 2025) [27]; (Mohaimin et al., 2025) [21].

For U.S. companies developing the tools that everyday users rely on, such as wallets, robo-advisors, and mobile trading apps, gaining predictive insights is invaluable. Liquidity-aware forecasting can enhance trade automation, provide real-time advisory features, and create more adaptive risk dashboards. These capabilities are not mere enhancements; they are essential, especially in a market that caters to a diverse audience, from urban power traders to rural first-time investors. As noted by Hossain et al. (2025) [12], the gaps in behavior and access are significant. Furthermore, Islam et al. (2025) [13] strongly advocate for the utilization of alternative data sources, like e-commerce trends, to develop models that can adapt to various financial realities.

## 5.4. Limitations

The LSTM model performed relatively well, and the liquidity features contributed more than we expected. However, there are several limitations in this study that we should not overlook. For starters, we only examined XRP paired with USDT, so applying these results to other cryptocurrencies, like Bitcoin, Ethereum, or lesser-known low-cap tokens, may not be valid. Each cryptocurrency has its unique dynamics. Additionally, we did not explore how changes in market sentiment, such as bull markets versus bear markets, could impact factors like volatility and sentiment. Developing models that can adjust to these shifts in real time could significantly improve our analyses. Moreover, we did not consider external factors, such as social media activity or on-chain events. While these signals might seem noisy, they have proven valuable in prior research. We also did not test for data stationarity or perform adequate data cleaning. Although the LSTM model can handle some degree of chaos, simpler models likely face difficulties. Finally, we did not take fraud into account. Issues like spoofing and wash trading can seriously distort liquidity signals. Research has shown that eliminating such noise can lead to much more reliable predictions.

Another important limitation of our modeling pipeline is the lack of regime detection. Cryptocurrency markets often go through cycles of bullish and bearish trends, each characterized by unique liquidity conditions and trader behaviors. Treating the entire dataset as if market behavior is uniform overlooks these structural shifts, potentially compromising the generalizability of our models. Additionally, while our models utilized high-frequency trading data, we did not perform rigorous tests for stationarity or address noise caused by market manipulation tactics such as spoofing or wash trading. These tactics can distort volume-based liquidity metrics and may have introduced bias into our input features. Furthermore, we did not incorporate mechanisms for online or continuous learning, which limits the model's ability to adapt to real-time events and changing market dynamics. This adaptability is crucial for high-frequency and intraday forecasting systems. Lastly, although we emphasized the predictive power of liquidity features, we did not establish quantitative benchmarks or operational thresholds that practitioners could use to monitor these indicators in production environments.

# 6. Future work

## 6.1. Broader cryptocurrency universe

This study shows that bringing liquidity into the mix can make machine-learning models a lot more effective at predicting crypto prices, at least in the case of XRP/USDT. But the real question is: does this approach hold up when applied to other cryptocurrencies? That's where future research comes in. It would be valuable to test this same modeling framework on a broader set of digital assets, such as Bitcoin, Ethereum, and Solana, to see if the improvements we saw here carry over. Each of these coins plays by slightly different rules. Bitcoin tends to have more consistent institutional backing and deeper liquidity, while altcoins like Solana can be a bit more volatile and prone to sudden speculative swings due to thinner order books. These differences aren't just quirks; they're tied to each token's underlying design, like how their networks function, what kinds of incentives exist in their ecosystems, and how exposed they are to systemic risks. While Bitcoin was originally introduced by Nakamoto (2008) as a decentralized alternative to fiat, the behavior of newer tokens is shaped by a much more complex set of forces [22].

Expanding this research to cover a range of coins wouldn't just test the model's robustness; it could also pave the way for more adaptable forecasting tools. Tools that aren't overly tailored to one asset, but instead can flex across different token types and market conditions. It would also open the door to interesting comparisons: how do liquidity indicators like VWAP or VVR interact with features like staking, DeFi use cases, or unique tokenomics? In the bigger picture, validating this approach across different assets could boost its usefulness for portfolio-level predictions, better risk management, and even cross-asset trading strategies. That makes it a potentially valuable tool not just for researchers but for traders, investors, and fintech platforms navigating the ever-evolving U.S. cryptocurrency landscape.

Future studies should not only apply the framework across different tokens but also examine how liquidity metrics interact with the market structures of these assets. For instance, Ethereum's transition to proof-of-stake may impact transaction volumes and VWAP dynamics differently than Bitcoin's UTXO model. Additionally, tokens associated with DeFi protocols or staking rewards may display price behaviors that are more influenced by liquidity lockups or farming incentives. Exploring these structurally diverse ecosystems would not only test the framework's robustness but also enhance our understanding of asset-specific liquidity sensitivities. This knowledge could assist in developing strategies for multi-token portfolios, where liquidity risk is unevenly distributed among holdings.

## 6.2. Additional feature sources

One exciting direction this research could take next is pulling in new kinds of data, especially those that tap into what people are thinking and doing in real-time. Social media platforms like Twitter, Reddit, and even Google Trends have become goldmines for gauging investor sentiment, which plays a big role in cryptocurrency markets where retail traders and herd behavior often drive price swings. If we bring in sentiment scores generated by natural language processing models, we might be able to catch those sudden surges or dips in the market that don't show up in price charts or trading volume alone (Al Montaser et al., 2025) [1].

There's also a lot of untapped potential in on-chain data. Things like the number of active wallet addresses, how much is being transacted, miner earnings, and how quickly tokens are moving around all give us a clearer picture of what's happening under the hood. These metrics act like the vital signs of a blockchain, pointing to healthy growth or early signs of strain. If we combine this kind of on-chain insight with our current liquidity-aware models, we might not only get more accurate predictions but also gain a deeper understanding of why prices move the way they do, especially for tokens whose value is tied to how they're used on the network.

To enhance interpretability and reduce the opacity of model behavior, we propose the incorporation of explainable AI (XAI) techniques into the modeling pipeline. Tools such as SHAP (SHapley Additive Explanations) and LIME (Local Interpretable Model-Agnostic Explanations) can provide detailed insights into how features like liquidity ratios, sentiment scores, and wallet activity influence model predictions. These tools are especially valuable in institutional settings, where decision-makers require not only accuracy but also transparency. Furthermore, integrating XAI methods could reveal hidden dependencies between on-chain data and liquidity signals, offering a more nuanced understanding of how both behavioral and structural market factors interact.

## 6.3. Advanced modeling techniques

As model designs keep advancing, it's worth exploring newer architectures like transformer-based models, think Temporal Fusion Transformer (TFT), for example. These models are especially good at picking up on long-term patterns and handling messy, uneven time series data with lots of variables in play. Unlike traditional recurrent networks, transformers use attention mechanisms, which basically means they can zero in on the most important parts of a sequence instead of treating everything equally. That makes them a solid fit for financial markets, where things can change fast and unpredictably. Looking ahead, there's also a lot of promise in blending models, say, combining LSTMs or transformers with tree-based methods like XGBoost or LightGBM.

These kinds of hybrid setups can shine by capturing both the flow of sequential data and the messy, nonlinear relationships that often pop up in financial datasets. Plus, they offer more transparency, helping us understand which features are driving predictions and how confident the model is, which is critical when real money is on the line (Jakir et al., 2023) [16]. And let's not forget about fraud detection. Mixing neural networks with rule-based models can build much stronger defenses, especially in high-stakes areas like credit risk. Sizan et al. (2025) also make a strong case for layered, hybrid approaches that bring together the best of both worlds, flexibility, and interpretability [28].

Future research could explore the development of hybrid architectures that combine different types of models and operate at various time scales. For example, a multi-resolution ensemble could consist of a transformer model focused on long-term macro cycles, paired with a tree-based model trained to analyze minute-level liquidity shifts. By aggregating forecasts from these complementary components, such a system may more effectively capture both structural and transient factors influencing price movements. Additionally, incorporating interpretable models like gradient boosting on top of deep learning embeddings could enhance both accuracy and auditability, thus bridging the gap between black-box and white-box forecasting in cryptocurrency environments.

## 6.4. Regime-aware and online learning

One area that deserves more attention is what's known as regime-aware modeling. Crypto markets don't behave the same all the time; they swing between bullish runs and bearish downturns, each with its quirks in terms of volatility, liquidity, and how traders react. If your model treats every market phase the same, it's going to miss a lot. That's where regime detection comes in. Tools like hidden Markov models or even something as straightforward as spotting volatility breakouts can help the system recognize what kind of market it's dealing with. From there, it can adjust, either by switching models or re-balancing its influence, so the forecasts stay relevant as conditions shift. Then there's the bonus of online learning. With methods like recurrent neural networks, streaming gradient descent, or adaptive boosting, your model doesn't just sit there; it keeps learning. It updates itself on the fly, which is a huge win in fast-moving environments like intraday or high-frequency trading. After all, in markets that can flip in minutes, even a slightly outdated model can drag down performance quickly (Mohaimin et al., 2025) [21].

A practical way forward involves using Hidden Markov Models (HMMs) or unsupervised clustering to classify regime types based on factors like volatility, volume, or spread compression. These labeled regimes can then inform downstream models or help choose between specialized submodels tailored for each type of regime. Additionally, implementing streaming data pipelines would enable the model to update incrementally rather than relying on batch retraining. This enhances adaptability, especially during major macroeconomic events or sudden regulatory announcements that can quickly change market dynamics, making static models ineffective.

## 6.5. Deployment considerations

At the end of the day, bringing forecasting models into real-world production isn't just about having strong predictions; it's about navigating a whole set of practical, often messy, challenges. One of the big ones is latency. If you're plugging these models into automated trading systems or high-frequency setups, even a slight delay can throw things off. Deep learning models like LSTMs or transformers are powerful, no doubt, but they can be heavy, and that computational load can slow down execution if you're aiming for real-time responsiveness. That's why future work needs to dig into things like model compression, quantization, and inference optimization, basically, anything that can help squeeze more performance out without sacrificing too much accuracy. Just as important is how easily these models can be integrated into the broader trading and risk infrastructure. Institutions aren't just looking for predictions; they need models that play nicely with their dashboards, compliance systems, and execution platforms. That means building APIs, creating live monitoring tools, and making sure everything is above board from a regulatory standpoint, especially in the U.S., where transparency and interpretability aren't just nice-to-haves anymore; they're starting to become legal requirements (Ray et al., 2025) [25].

Interestingly, Rana et al. (2025) also point out that financial institutions are leaning toward ML dashboards that strike a balance, strong predictive power but with clear visibility into how the models work [23]. Rahman et al. (2025) go even further, highlighting how blockchain and digital ledger tech could help bring more accountability and traceability to real-time AI systems [24]. And in the backdrop of all this, there's the financial instability of some crypto firms, which raises the stakes. Sizan et al. (2025) make a compelling case that machine learning–based bankruptcy prediction tools could serve as early warning signs, flagging risky entities that might quietly destabilize token value or drain liquidity before anyone sees it coming [29].

Future implementations should define and validate actionable benchmarks for liquidity indicators such as VWAP (Volume Weighted Average Price) and VVR (Volume-Volume Ratio). For instance, traders could receive alerts when the VWAP deviates from the spot price by a certain threshold or when the VVR exceeds its rolling average by a predefined margin. Incorporating these triggers into real-time dashboards would enhance their utility and make the insights from our models more actionable. Additionally, establishing standardized explainability metrics, like the average SHAP (Shapley Additive Explanations) contribution for liquidity inputs, could assist regulators in assessing fairness and robustness in institutional contexts, especially as transparency becomes increasingly important.

# 7. Conclusion

In this study, we set out to build smarter machine learning models for predicting cryptocurrency prices, zeroing in on the XRP/USDT trading pair within the U.S. digital asset market. A big part of what drove this work was the realization that most forecasting models tend to sideline a pretty important factor: liquidity. So, we decided to bring it into focus using two specific metrics, the Volume-to-Volatility Ratio (VVR) and the Volume-Weighted Average Price (VWAP). These helped us dig deeper into the market's structure, beyond just price and volume. We tested a mix of models: Linear Regression, Random Forest, XGBoost, and LSTM, and found that LSTM consistently came out ahead in both accuracy and stability. When we added the liquidity indicators, performance improved across the board, which exemplified how essential those features are, not just in theory, but in practice. The LSTM model showed the strongest results, achieving the highest $R^2$ score and the lowest prediction error. That confirmed our hunch: models that understand sequences and time-based patterns are well-suited for the complexities of cryptocurrency price behavior.

But this isn't just about building better models. The results carry weight for traders, fintech developers, and researchers alike. For algorithmic traders, integrating real-time liquidity signals could mean more precise execution and better risk control. For the academic and ML communities, this work pushes the case for making liquidity features a standard part of crypto-focused prediction research. Ultimately, what we're aiming for is the kind of forecasting system that can handle different coins, shifting conditions, and the wild swings that come with cryptocurrency markets. Of course, our scope was limited, we focused on one trading pair and didn't include things like market sentiment or on-chain data. But that leaves a clear path forward. Expanding the asset pool, layering in multiple types of data, and experimenting with more advanced models like transformers or regime-aware systems could take this work to the next level. As the digital asset space continues to evolve, so should our models, grounded in real-world signals, flexible across markets, and built with both performance and interpretability in mind.